\documentclass[conference]{IEEEtran}
\IEEEoverridecommandlockouts
\usepackage{amsmath,amssymb,amsfonts}
\usepackage{graphicx}
\usepackage{textcomp}
\usepackage{enumitem}
\usepackage{balance} 
\usepackage{cite}
\usepackage{xcolor}
\usepackage[numbers]{natbib}
\usepackage[bookmarks=true]{hyperref}

\usepackage{comment}
\usepackage{hhline}
\usepackage{caption}
\usepackage{subcaption}
\usepackage{amsmath}
\usepackage{tikz}
\usetikzlibrary{positioning}
\usepackage[export]{adjustbox}
\usepackage{pgfplots}
\usepackage{pgfplotstable}
\pgfplotsset{compat=1.7}
\usepgfplotslibrary{groupplots}
\usepackage{multirow}
\def\BibTeX{{\rm B\kern-.05em{\sc i\kern-.025em b}\kern-.08em
    T\kern-.1667em\lower.7ex\hbox{E}\kern-.125emX}}

\newcommand{\newlineauthors}{%
  \end{@IEEEauthorhalign}\hfill\mbox{}\par
  \mbox{}\hfill\begin{@IEEEauthorhalign}
}
\makeatother

\makeatletter

\let\old@ps@IEEEtitlepagestyle\ps@IEEEtitlepagestyle
\def\confheader#1{%
    \def\ps@IEEEtitlepagestyle{%
        \old@ps@IEEEtitlepagestyle%
        \def\@oddhead{\strut\hfill#1\hfill\strut}%
        \def\@evenhead{\strut\hfill#1\hfill\strut}%
    }%
    \ps@headings%
}
\makeatother

\usepackage{tikz}
\newcommand\copyrighttext{%
  \footnotesize \textcopyright 2023 IEEE. Personal use of this material is permitted.
  Permission from IEEE must be obtained for all other uses, in any current or future
  media, including reprinting/republishing this material for advertising or promotional
  purposes, creating new collective works, for resale or redistribution to servers or
  lists, or reuse of any copyrighted component of this work in other works.}
\newcommand\copyrightnotice{%
\begin{tikzpicture}[remember picture,overlay]
\node[anchor=south,yshift=10pt] at (current page.south) {\fbox{\parbox{\dimexpr\textwidth-\fboxsep-\fboxrule\relax}{\copyrighttext}}};
\end{tikzpicture}%
}

\begin{document}

\title{``How Did They Come Across?'' Lessons Learned from Continuous Affective Ratings
}

\author{%
  \IEEEauthorblockN{%
    \parbox{\linewidth}{\centering
      Maria Teresa Parreira\IEEEauthorrefmark{1},
      Michael J. Sack\IEEEauthorrefmark{1},
      Hifza Javed\IEEEauthorrefmark{2},
      Nawid Jamali\IEEEauthorrefmark{2}, and
      Malte Jung\IEEEauthorrefmark{1}
    }%
  }%
  \IEEEauthorblockA{%
    \IEEEauthorrefmark{1}Cornell University, 
    \IEEEauthorrefmark{2}Honda Research Institute USA, Inc.
  }%
}
\maketitle
\copyrightnotice

\begin{abstract}

Social distance, or perception of the other, is recognized as a dynamic dimension of an interaction, but yet to be widely explored or understood. Through CORAE, a novel web-based open-source tool for \textit{COntinuous Retrospective Affect Evaluation}, we collected retrospective ratings of interpersonal perceptions between 12 participant dyads. In this work, we explore how different aspects of these interactions reflect on the ratings collected, through a discourse analysis of individual and social behavior of the interactants. We found that different events observed in the ratings can be mapped to complex interaction phenomena, shedding light on relevant interaction features that may play a role in interpersonal understanding and grounding. This paves the way for better, more seamless human-robot interactions, where affect is interpreted as highly dynamic and contingent on interaction history.
\end{abstract}


\IEEEpeerreviewmaketitle

\section{Introduction}

Affect is a dynamic phenomenon characterized by the temporal evolving nature of observable behavior, subjective experience, and physiology \cite{kuppens2017emotion}. While previous research has made significant progress in capturing affect data over time across domains such as physiology \cite{healey2005detecting}, subjective experience \cite{ruef2007continuous, csikszentmihalyi2014validity}, and observable behavior \cite{jung2016thin, coan2007specific}, there remains a gap in our understanding of how affective perceptions of others dynamically develop over time.

Retrospective analysis is an established method for collecting continuous affect data \cite{cowie2007feeltrace,cowie2013gtrace,melhart2019pagan,lopes2017ranktrace}. Typically, this approach involves video-recording participants during an event that elicits emotions; later, participants are asked to continuously rate their feelings while watching the recorded video \cite{ruef2007continuous}. This method relies on the phenomenon that individuals often re-experience emotions when reliving a situation \cite{gottman1985valid}. With CORAE, a novel browser-based tool for \textit{\textbf{CO}ntinuous \textbf{R}etrospective \textbf{A}ffect \textbf{E}valuation}, we build upon existing work by introducing an approach and a tool that enable researchers to collect continuous affect data of interpersonal perceptions.

This intuitive tool allows participants to retrospectively rank how another interactant came across immediately following an interaction, thus allowing us to capture interpersonal affective perceptions rather than feelings or affective state inferences. Our approach is grounded in a behavioral ecology perspective of emotion, which posits that emotional expressions are primarily social tools used to influence and learn about others \cite{fridlund2014human, crivelli2019inside, van2010emerging}. In line with this perspective, rather than capturing valence when rating interactions, we focus on a dimension of approach and withdrawal, i.e., the degree to which behavior is seen as increasing or decreasing social distance \cite{jung2017affective, caffi1994toward, andersen1996principles}.

 \begin{figure}[t] 
    \centering
    \includegraphics[width=0.43\textwidth]{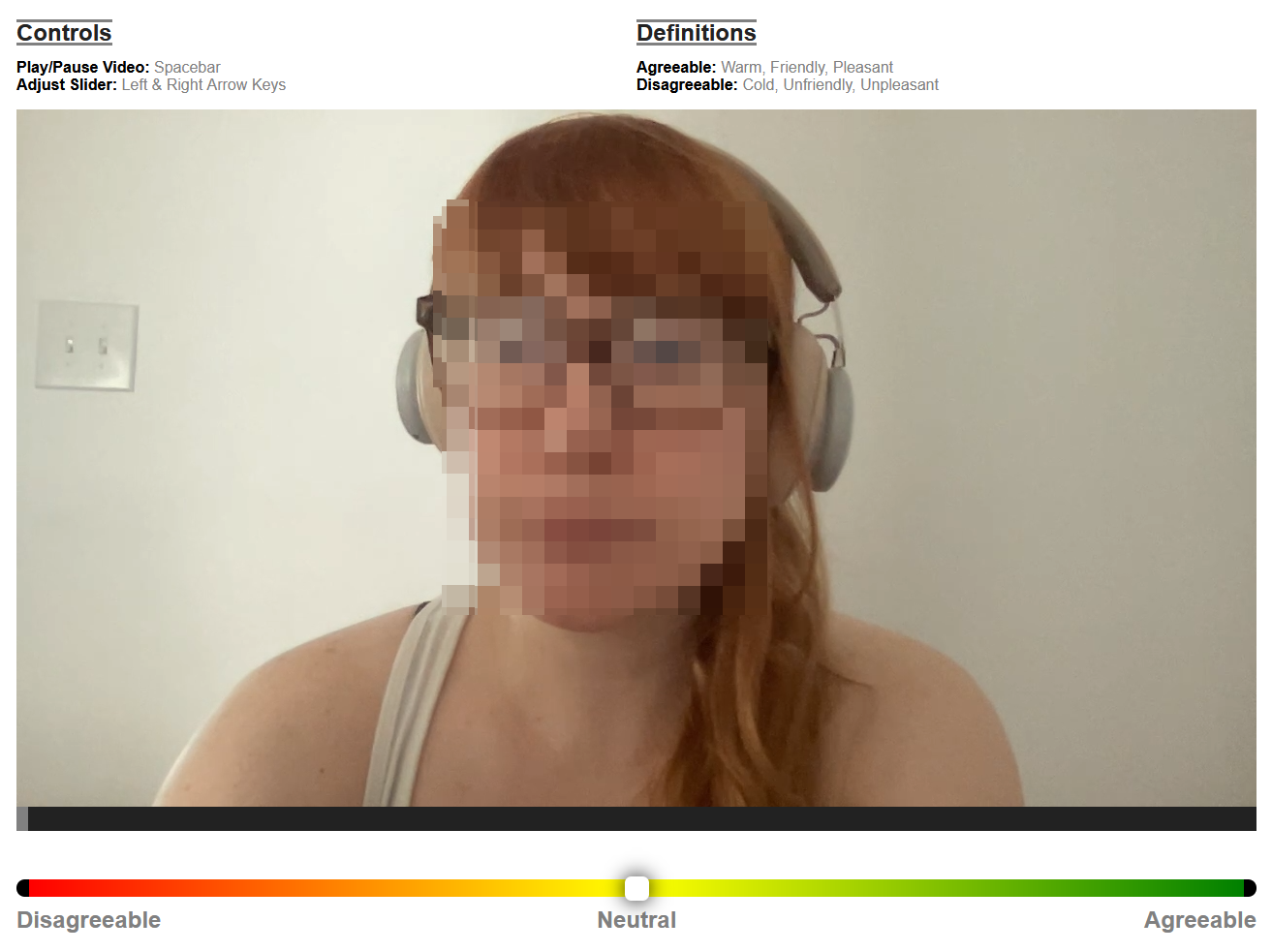}
    \caption{Annotation dashboard for CORAE. After interacting, participants are asked to retrospectively evaluate how the other person came across. They review only the video from the other participant and audio of both, playing simultaneously.}
    \label{fig:frontend}
\end{figure}

In order to validate the platform, we collected retrospective ratings from a total of 24 participants, in 12 dyadic interaction sessions \cite{2023CORAE}. In this work, we re-analyse the data collected, aiming to extract a deeper understanding of how the ratings reflect specific interaction phenomena.
\section{CORAE}

We provide below a brief description of CORAE, to facilitate the understanding of the interpersonal perception data used in this study.
The CORAE platform (related to the Latin word for ``heart'') \cite{2023CORAE} enables individuals to intuitively evaluate how a person's behavior is interpreted emotionally during interactions. It is an open-source tool that can be found in \href{https://corae.org}{corae.org}.

\subsection{Design}
\label{subsec:design}
Building upon existing tools, we sought to address features and applications under-served by the broader design space and incorporate those most aligned with an intuitive annotation experience. y training or resources to use them. 

\begin{description}[align=left, leftmargin=0em, labelsep=0.2em, font=\textbf, itemsep=0em,parsep=0.3em]

\item[Layout: ]
We designed CORAE to be intuitive and visually minimal (see \autoref{fig:frontend}) in its presentation. CORAE is deployed in a web browser with the central focus being a video of one's co-interactant, staged for annotation. This is to ensure participants are not distracted by their own image on the screen nor by other visual elements on the platform. Brief instructions above the video player describe keybindings to control the annotation dashboard (\textit{Spacebar} to toggle playback and \textit{Left and Right Arrows} to control the slider), as well as a brief description of the terms used for measuring interpersonal perception.

 An unobtrusive progress bar is displayed below the video player to inform participants what proportion remains of their evaluation. Finally, below the video player is displayed the annotation slider. A color gradient (from red to green) enables participants to more intuitively understand the meaning of each of the rating terms. The annotation bar is bounded and discretized (a total of 15 points, from $-7$ --- \textbf{Disagreeable} --- to $+7$ --- \textbf{Agreeable}). Participants may only change their rating during video playback and are constrained by the platform to do it ``continuously" (i.e., they cannot instantaneously change the rating from \textit{Neutral} ($0$) to \textit{Agreeable} ($7$), but rather adjust to each value in sequence).  

\item[Intuitiveness: ]
By mitigating the need to train annotators on CORAE's use, our platform facilitates continuous retrospective annotation immediately following an interaction. The immediacy of this evaluation allows for stronger salience of affect compared to a delayed approach. 

\item[Affective Dimension: ]
 The complexity of annotating two or more concurrent affective dimensions imposes a nontrivial cognitive demand upon the annotator \cite{cowie2013gtrace}. This demand, in turn, may diminish the salience of user annotation along both dimensions. Similar to PAGAN \cite{melhart2019pagan} and others, our design aims to capture affect along a single dimension.

\item[Distributed Participation: ]
Whereas existing solutions tend to rely heavily upon the collocation of researchers and participants for in-person data collection, we found this approach to be unnecessarily restrictive and rather limiting to the potential recruitment of more diverse populations. Conversely, remote studies are unable to guarantee consistency for factors such as participant system specifications and environmental distractors. We acknowledge the value afforded by both in-person and remote study formats and sought to develop a tool capable of facilitating either. To this end, CORAE may be deployed locally for in-person sessions as well as remotely for distributed participation.

\end{description}

\subsection{Functionality}
\label{subsec:functionality}

We focused our early development efforts on creating an intuitive and seamless annotation experience for the user. To this end, we withheld several planned features with the intention of their inclusion in a subsequent release of the platform (Section \ref{subsec:release}). When we refer to CORAE and its functionality in this section, it is in reference to build~0.15a unless otherwise noted. The features in this build were those we deemed critical for CORAE's experimental validation and ongoing user testing.

\begin{description}[align=left, leftmargin=0em, labelsep=0.2em, font=\textbf, itemsep=0em,parsep=0.3em]
\item[Annotation Dashboard: ]
 A unique URL is generated for each participant to access their instance of CORAE's annotation dashboard (Fig. \ref{fig:frontend}). Upon accessing their instance, participants are, by default, prompted to enter an identifier which is then logged by the platform and associated with their session. 
 
 In terms of interaction, the dashboard affords two primary actions: slider adjustment and playback control. Annotators may indicate their affect rating by adjusting the slider using \textit{Left and Right Arrow} during playback. This affect rating by default is indicated using a continuous 15-point scale but may be changed in the project template to suit any granularity.
 
\item[Data Logging: ]
Data is logged for a session in two ways: (1) by default, the mode for data logging is set to predetermined intervals of one second, which may be adjusted to any granularity; and (2) to ensure accuracy in the annotation method, CORAE also logs data whenever a change in the rating occurs. Associated data points are the slider position (rating), time code, and video frame (in the format \textit{``SliderNumericalPosition": ``Hours:Minutes:Seconds:VideoFrame"}), which are logged in a JSON file. Given that video is recorded at a rate of 30 frames-per-second, this allows for a resolution of up to $1/30$ s in the annotated data streams.
\end{description}

\section{User Study}
\label{sec:study}

In prior work, we tested a use case for CORAE in the form of an experimental study on interpersonal dynamics during dyadic interactions. We briefly describe the study design below.

\subsection{Experimental Procedure}

Participants were recruited through Prolific\footnote{\url{https://www.prolific.co/}}. The study took place fully online. Before scheduling their slot, each participant read and signed a consent form. At the scheduled time for the experiment, both participants received a link to a call on Zencastr\footnote{\url{https://zencastr.com/}}, a video call platform that allows for high-quality recording of each video and audio stream separately. Participants read task instructions, including a description of the discussion topic (Reasons for Poverty task \cite{shek2002reasons}, detailed in Section~\ref{sec:reasons-for-poverty}). After this, participants were recorded while interacting to solve the task. When they reached an agreement, or after 10 minutes of discussion, participants were asked to stop discussing and fill out a survey. This survey collected demographic data, as well as measures of interpersonal affect. In the meantime, the researcher downloaded the data streams, merged the video stream with audio streams from both participants and uploaded them to CORAE. Each participant was then distributed a unique URL which opened an instance of CORAE's annotation platform in their browser. Participants were each presented with a video of their discussion partner were asked to continuously rate how their partner came across moment-to-moment. Once finished, participants were instructed to download the annotation file and upload it onto an encrypted database. Finally, after completing an exit survey, participants were compensated for their participation with US\$14, through Prolific. 

\subsection{Reasons for Poverty task}
\label{sec:reasons-for-poverty}

To evaluate our tool, we needed a task that could elicit a broad range of emotions. We used a modified version of the \textit{Reasons for Poverty} task \cite{shek2002reasons}. The task requires participants to rank order a list of ``reasons for poverty'' according to their ``accuracy''. Half of the items follow a reasoning that sees the source of poverty in peoples' situation, i.e., their circumstances, whereas the other half follows a reasoning that sees the source of poverty in peoples' disposition, i.e., their personality. By strategically recruiting participants with opposing beliefs about poverty, we aimed to elicit an emotionally engaging interaction. We used the following instructions:

\textit{You and the other participant must come to an agreement as to a rank of the 5 most relevant causes of poverty in order of the accuracy of each statement. The cause of poverty that is evaluated as being most accurate will be ranked as 1st, and the one that is evaluated as least accurate will be ranked as 5th.}
\textit{
\begin{itemize}
\item Poor people lack the ability to manage money. 
\item Poor people waste their money on inappropriate items. 
\item Poor people do not actively seek to improve their lives. 
\item Poor people lack talents and abilities. 
\item Poor people are exploited by the rich. 
\item The society lacks justice. 
\item Distribution of wealth in society is uneven. 
\item Poor people lack opportunities because they live in poor families. 
\item Poor people live in places where there are not many opportunities. 
\item Poor people have encountered personal misfortunes, which limit their opportunities. 
\item Poor people are discriminated against in society. 
\item Poor people have bad fate. 
\item Poor people lack luck. 
\end{itemize}}

Participants were given a maximum of 10 minutes to discuss, to prevent individuals from getting disengaged when reviewing their discussion on CORAE.

\subsection{Participants}

Participants were recruited through Prolific. To elicit disagreement during the interactions, participants were selected according to their political leaning (one conservative- and one liberal-leaning). Other recruitment criteria were proficiency in English and a computer device with a functioning camera and microphone.

\subsection{Analysis}

To get insights on the interaction features that play a role on observed rating phenomena, we selected 5 moments of interest based on the \textbf{Cumulative Interpersonal Rating (CIR)} curves. These curves are obtained by the cumulative sum of the ratings for each participant (e.g., see Figure \ref{fig:moment1}). We looked for instants of synchronicity or opposition in the ratings. For the analysis, we follow principles of ethnomethodology to review the interaction moment selected, but also analyze the preceding and following moments to provide context for the potential reasons and consequences of such interaction phenomena.

\section{Results}
\label{sec:results}


Below, we describe each interaction event selected for analysis. We include a description of the observed CIR curve, participant demographics, a description of the interaction phenomena observed through video analysis and, finally, observations from the video analysis of prior and posterior moments of that session, to contextualize the behaviors in a broader lens.

\subsection{Event - Sudden Drop}
\label{sec:moment-1}

\subsubsection{CIR Curve}

\begin{figure}
    \centering
    \includegraphics[width=0.8\linewidth]{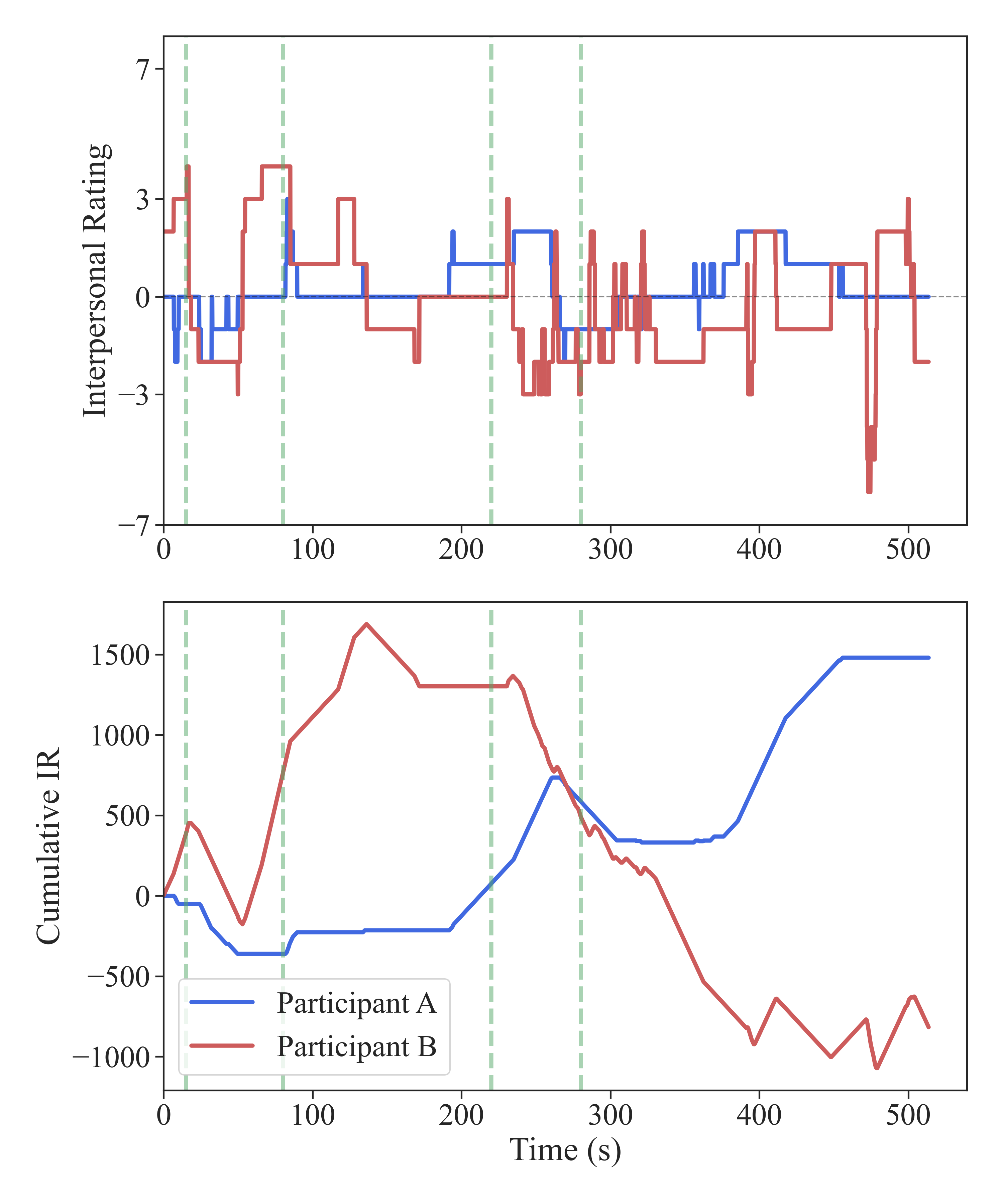}
    \caption{Ratings (top) and CIR curves (bottom) for each participant in session S1, and phenomena analyzed (between green lines), corresponding to seconds $15-80$ and $220-280$ of that interaction.}
    \label{fig:moment1}
\end{figure}

We selected an event from session S1 (seconds $15-80$, see Figure \ref{fig:moment1}). Both ratings decrease, then rebound almost simultaneously, very early into the interaction.

\subsubsection{Demographics}
Participants in this session, aged 31 (A) and 46 (B), both identified as Male and reported native English proficiency.

\begin{figure} 
    \centering
    \includegraphics[width=0.7\linewidth]{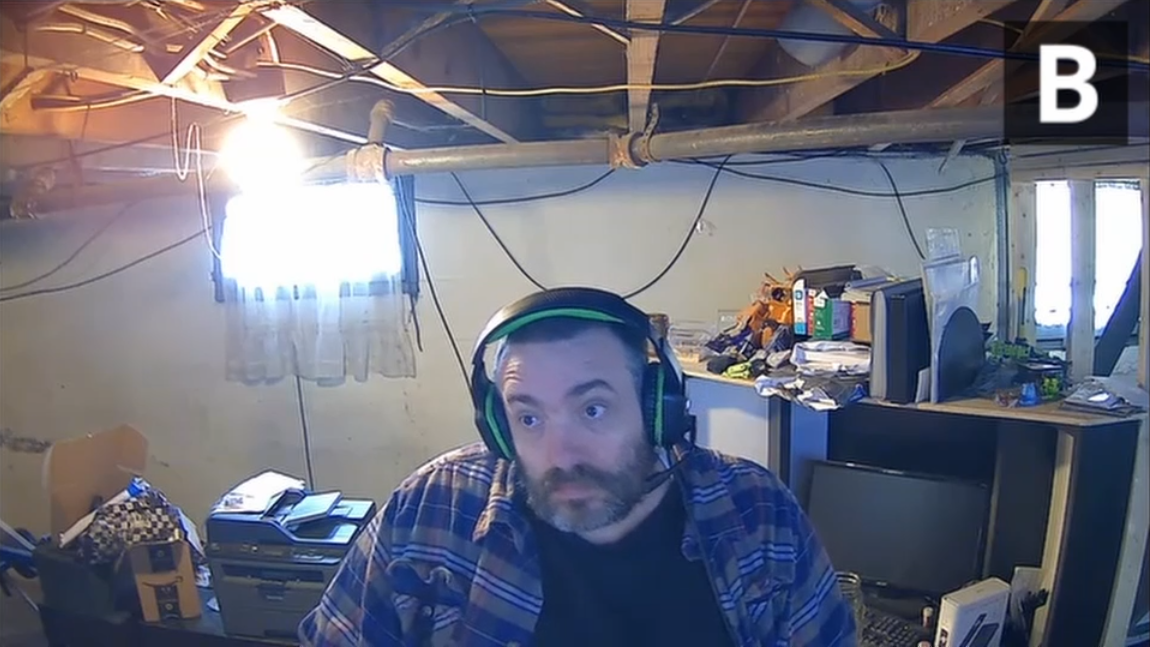}
    \caption{PB emits non-verbal backchannel in disagreement (i.e. raised eyebrows with lip-corner depressor), corresponding to seconds $29-33$ of that interaction.}
    \label{fig:45-frown}
\end{figure}

\subsubsection{Event}
Early on in the task, lacking insight into their discussion partner's ideology, PA expresses an opinion with a clear ideological leaning. PB listens without interrupting, but appears generally disagreeable, sometimes looking away while raising their eyebrows with a frown (see Figure \ref{fig:45-frown}). During this time, PA verbally narrates through the task list, and does not ask for PB's opinion. As PA continues, PB provides no feedback, non-verbal or otherwise. Following $65$ seconds of unbroken monologue, PA asks PB for their opinion on an ordering for the task list.

\subsubsection{Context}
Given that this event occurs at the beginning of the session, there is no context that precedes it. The featured interaction is the first exchange between the participants, who were previously unaware of their opposing ideology. Following this event, PA demonstrates responsive listening (e.g. verbal backchanneling and nodding) toward PB when PB eventually shares their own opposing perspective.

\subsection{Event - Opposing trends}
\label{sec:moment-2}

\subsubsection{CIR Curve}
We selected a second event from session S1 (seconds $220-280$, see Figure \ref{fig:moment1}). CIR curves show opposite trends, with one increasing and one decreasing in value, almost with the same slope.

\subsubsection{Event}
In a second clip from the same session, PB begins asserting their opinions in clear opposition to those established by PA. The CIR for this period depicts a sharp and sudden decrease in rating of PB's perception toward PA. Further, PB remarks at the perceived cliche of their argument, apparently preempting its defense. PA emits both verbal and non-verbal backchannels such as nodding and utterances of acknowledgment such as "okay" and "mhm".

\subsubsection{Context}
Following this event, PA presents an argument as to why they do not entirely agree with the opinion of PB. Both participants' ratings decrease during this period, however, PA's ratings moderately rebound to a neutral state at around second $300$. Further, we observe several successive rebounds in PB's CIR curve at $400$ seconds, coinciding with a repeated shift in conversational control between both participants. Despite this, PB's CIR concludes firmly in the negative. This suggests that PB's perception of PA was unable to recover following this event.

\subsection{Event - CIRs cross}
\label{sec:moment-4}

\subsubsection{CIR Curve}

\begin{figure} 
    \centering
    \includegraphics[width=0.8\linewidth]{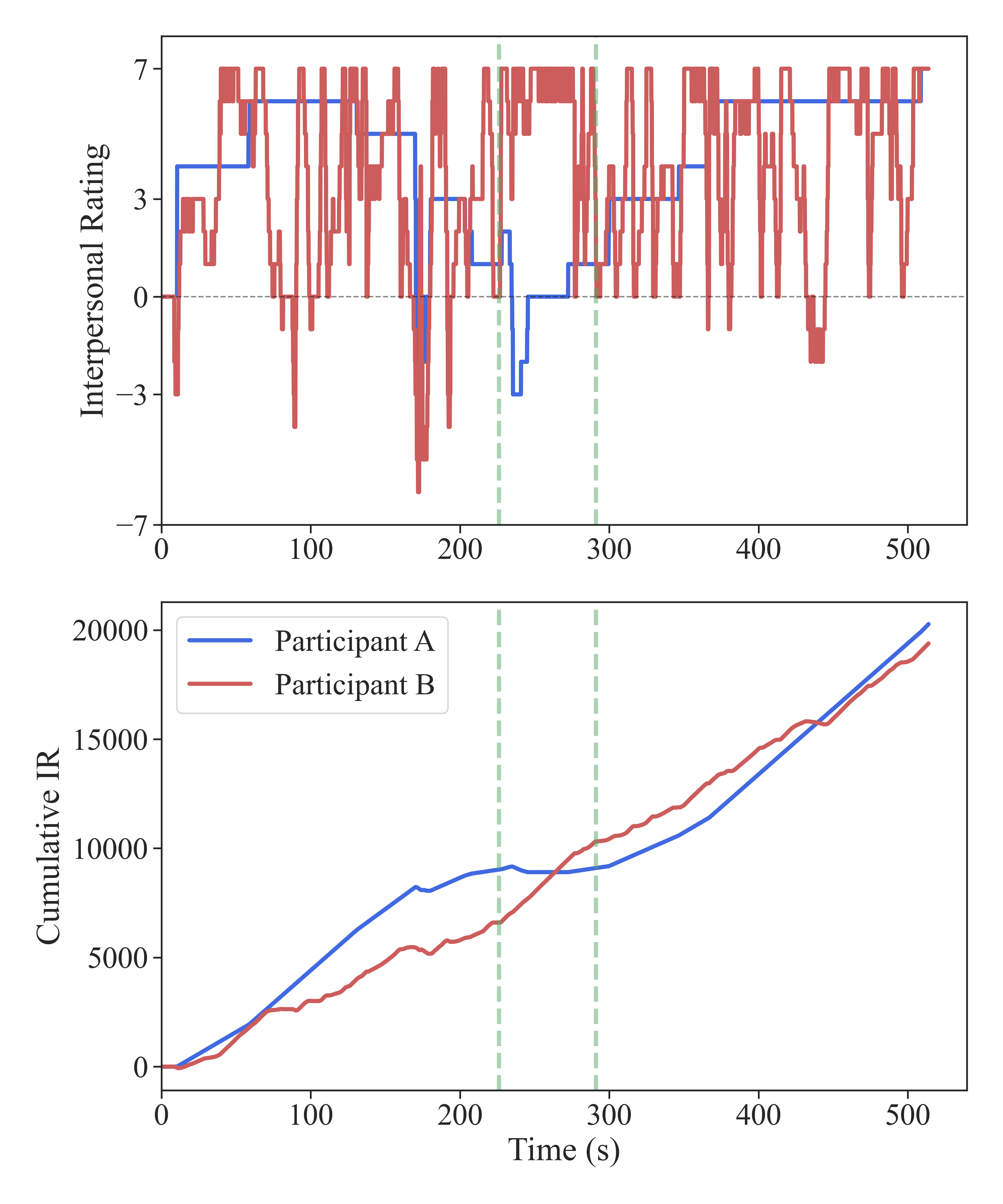}
    \caption{Ratings (top) and CIR curves (bottom) for each participant in session S2, and phenomenon analyzed (between green lines), corresponding to seconds $226-291$ of the interaction.}
    \label{fig:moment2}
\end{figure}

We selected an event from session S2 (seconds $226-291$, see Figure \ref{fig:moment2}). One participant's rating plateaus, while the other increases steadily, leading the curves to cross.

\subsubsection{Demographics}
Participants in this session, aged 27 (A) and 61 (B), both identified as Female and reported either native English proficiency (A) or high proficiency (B).

\subsubsection{Event}

In this selected clip, PB asserts disagreement with one of PA's opinion and rebuts, making clear a difference in ideology. Verbal interjections by PA go ignored by PB, who speaks over PA. PA appears visibly discouraged by each interruption, emitting less frequent backchannels compared to exchanges both before and after this event. 

\subsubsection{Context}
The interruptions of PA by PB discussed above are uncharacteristic of the surrounding discussion. In the pre-clip period, both participants seamlessly engage in equitable turn-taking as they work through the task list, politely yielding conversational control to each other when warranted. During this period, participants regularly express their agreement and attention through both verbal and non-verbal backchannels, remaining silent as the other speaks. The shared perception of these exchanges is reflected by a mutual positive trend in CIR leading up to the event featured in the selected clip. This trend continues in the post-clip period, as the participants reconcile their conflicting opinions and once again find common ground.

\subsection{Event - Plateaus and synchronicity}
\label{sec:moment-5}

\subsubsection{CIR Curve}

\begin{figure} [h]
    \centering
    \includegraphics[width=0.8\linewidth]{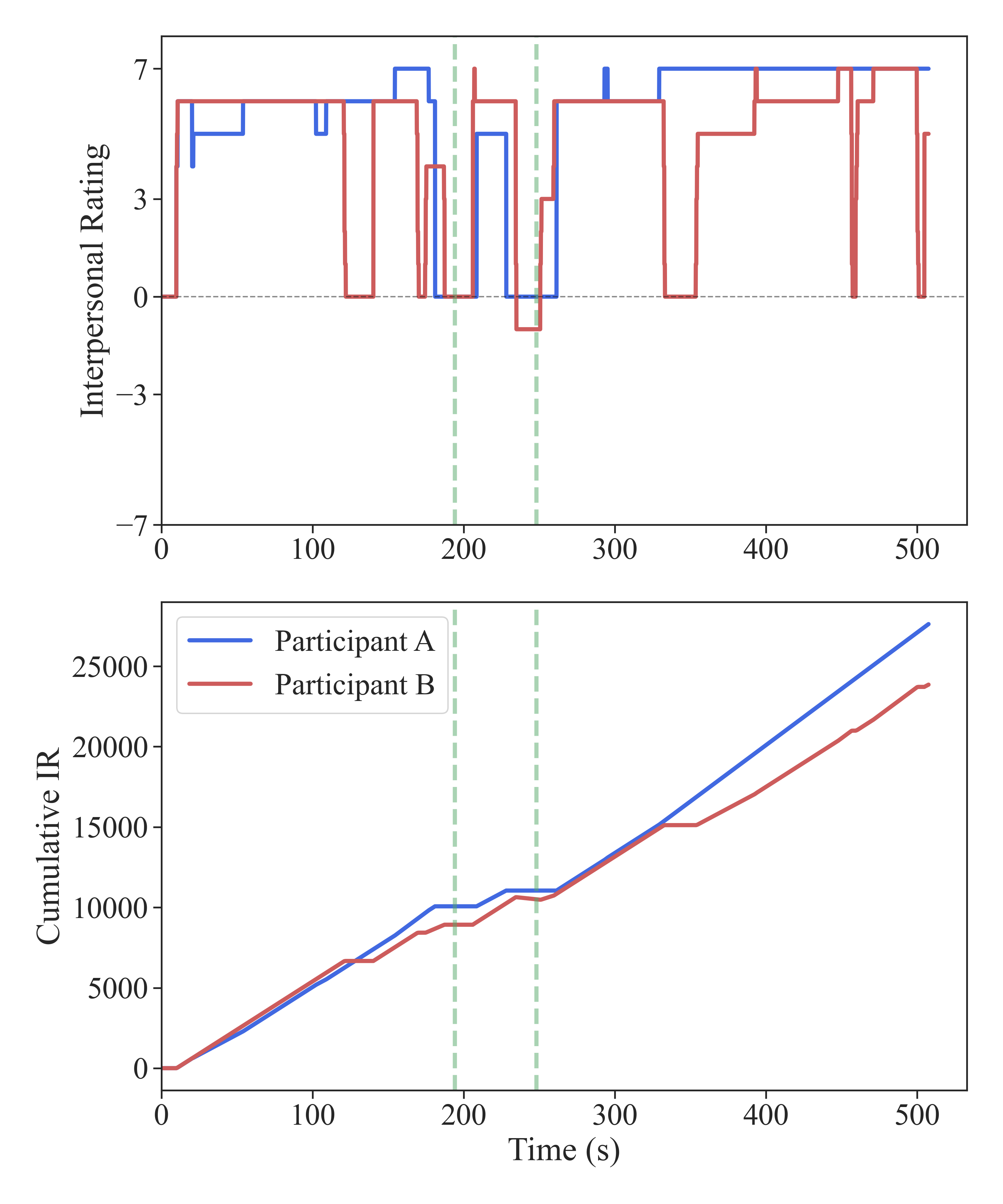}
    \caption{Ratings (top) and CIR curves (bottom) for each participant in session S3, and phenomenon analyzed (between green lines), corresponding to seconds $194-248$ of the interaction.}
    \label{fig:moment3}
\end{figure}

We selected an event from session S3 (seconds $194-248$, see Figure \ref{fig:moment3}). Both CIR curves somewhat plateau. From the IR curves, we can see synchronicity in the ratings of both participants (in spite of these annotation sessions taking place completely independently).

\subsubsection{Demographics}
Participants in this session, aged 39 (A) and 61 (B), both identified as Female and reported native English proficiency.

\subsubsection{Event}
\begin{figure} [h]
    \centering
    \includegraphics[width=0.7\linewidth]{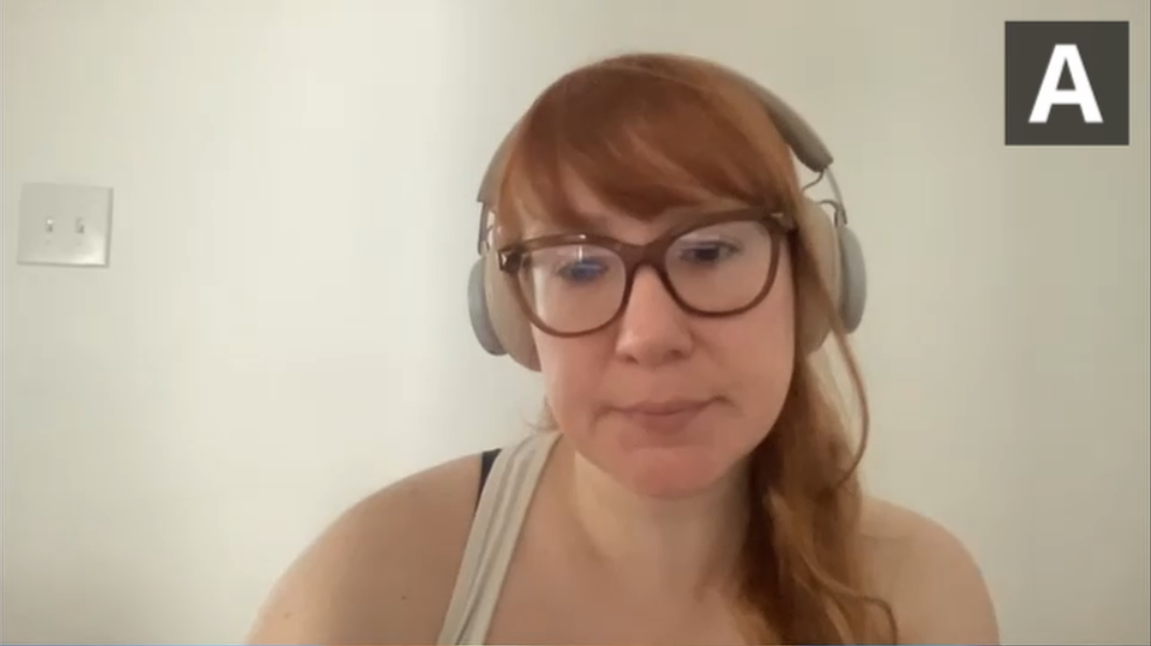}
    \caption{PA emits non-verbal backchannel (i.e. tilted head with lip pursing/tightening), corresponding to seconds $205-231$ of that interaction.}
    \label{fig:7879-purse}
\end{figure}

The event featured in this selected clip occurs following the participants' agreement on the ordering of several task items, reflected in a positive synchronicity depicted in the IR curve toward the beginning of this period. Having already agreed upon three of the five required items for the task, PB proposes a contender for position four and immediately moves to justify their proposal. As PB describes their reasoning, the frequency of both verbal and non-verbal backchannels emitted by PA drops significantly. PA becomes visibly more tense with heavily restricted body movement, in addition to a prolonged lip pursing across the duration of the event. During this period, PA emits sporadic non-verbal backchannels in the form of weak head nods and head tilts (see Figure \ref{fig:7879-purse}).

\subsubsection{Context}
Preceding this event, PA and PB are able to find common ground and agree strongly on several points early on. Notably, PA asserts one reason for poverty as a strong contender but hedges indecision because of a perceived "redundancy" with a previously agreed-upon item. This context is helpful for understanding PA's apparent confusion by PB's rebuttal, as well as the plateau in the CIR curve observed during this event. Only once during their discussion does either participant rate the other negatively (see Figure \ref{fig:moment3}), instead opting for neutral ratings during instances of disagreement. Crucially, the participants are able to reconcile their difference in opinion following this event in addition to the conflict that precedes it. This is illustrated by a strong synchronicity in the IR and CIR curves of both participants.

Following this event, PA's IR curve remains fairly consistent, trending positive for the remainder of the interaction (see Figure \ref{fig:moment3}). PB's IR curve oscillates more dynamically in response to later exchanges, though, never again falling below the neutral position.
\section{Discussion}
\label{sec:discussion}

This work dives into one of the use cases of CORAE, a novel approach to the collection of interpersonal data. In prior work \cite{2023CORAE}, we provided evidence that the tool is both accurate and intuitive/easy to use.
Through a cross-reference qualitative analysis of interaction moments and the interpersonal ratings curves, here we obtain insights into individuals' ability to continuously recall affective perceptions of their interaction participant. We also unveil the intricate dynamics of interpersonal perception, with complex and multimodal phenomena, including verbal and non-verbal behaviors, memory of past interaction moments, and regions of interpersonal perception synchronicity. The behavior observations, mapped to quantitative data from the interpersonal ratings, shed light on interaction dynamics and the nuanced nature of time-dependent perceptions of the other.

Crucially, we note that the retrospective analysis provides an opportunity to add different input modalities to the interaction, exposing behaviors from the other interactant that might have otherwise have been missed (e.g., Event \ref{sec:moment-1}), where PA was reading from the task prompt; when rating, PA gets to observe PB's non-verbal disagreeable behavior, which might have changed its perception of the interaction). This is a limitation to consider when analyzing the data exported from the retrospective analysis tool.

CORAE was developed to be intuitive, cutting the need for training sessions and thus increasing the potential for capturing affective states continuously over time. The platform allows for the collection of high-resolution data, with participants changing ratings often and across a broad range of values of interpersonal distance. CORAE thus constitutes a valuable tool that is also easy to customize to other interactions or experimental contexts. 

Our findings demonstrate the value of using continuous affect rating methods to capture the temporal dynamics of affect and emotion in social interactions. By allowing participants to rate their affective experience both continuously and retrospectively, our tool provides a more fine-grained understanding of the emotional experiences of individuals throughout an interaction. This can help researchers to better understand how affective experiences influence behavior and how behavior, in turn, shapes affective experiences \cite{mettallinou2013review}. Additionally, the continuous stream of retrospective annotation data offers potential for machine learning applications, with predictive affect systems playing an important role in the design of human-robot interactions.

\section{Conclusion}
\label{sec:conclusion}

In this work, we mapped individual and social behavior observations to quantitative data, obtained through interpersonal ratings collected with CORAE. We noted multimodal and composite phenomena, with impacts on the interaction that spread across time.
Our work contributes to the growing body of research on affective computing and provides CORAE as a valuable tool for investigating the temporal dynamics of affect and emotion in social interactions. We believe that this tool has the potential to shed light on the complex and nuanced nature of human emotional experiences and inform the development of more effective interventions and technologies for improving human-robot social interactions.

\section*{Acknowledgments}
This work was supported by Honda Research Institute USA, Inc..

\small
\bibliographystyle{abbrvnat}
\balance
\bibliography{bibliography.bib}

\end{document}